# A Novel Shape Guided Transformer Network for Instance Segmentation in Remote Sensing Images

Dawen Yu and Shunping Ji, *Senior Member, IEEE*

*Abstract*—Instance segmentation performance in remote sensing images (RSIs) is significantly affected by two issues: how to extract accurate boundaries of objects from remote imaging through the dynamic atmosphere, and how to integrate the mutual information of related object instances scattered over a vast spatial region. In this study, we propose a novel Shape Guided Transformer Network (SGTN) to accurately extract objects at the instance level. Inspired by the global contextual modeling capacity of the self-attention mechanism, we propose an effective transformer encoder termed LSwin, which incorporates vertical and horizontal 1D global self-attention mechanisms to obtain better global-perception capacity for RSIs than the popular local-shifted-window based Swin Transformer. To achieve accurate instance mask segmentation, we introduce a shape guidance module (SGM) to emphasize the object boundary and shape information. The combination of SGM, which emphasizes the local detail information, and LSwin, which focuses on the global context relationships, achieve excellent RSI instance segmentation. Their effectiveness was validated through comprehensive ablation experiments. Especially, LSwin is proved better than the popular ResNet and Swin transformer encoder at the same level of efficiency. Compared to other instance segmentation methods, our SGTN achieves the highest average precision (AP) scores on two single-class public datasets (WHU dataset and BITCC dataset) and a multi-class public dataset (NWPU VHR-10 dataset). Code will be available at http://gpcv.whu.edu.cn/data/.

## I. INTRODUCTION

INSTANCE segmentation aims to simultaneously detect objects of interest and delineate their foreground regions at a detailed level [1-3]. The earliest deep learning (DL) based instance segmentation method, SDS (Simultaneous Detection and Segmentation), was proposed by Hariharan et al. in 2014 [4]. Since then, instance segmentation methods have rapidly advanced. Today, DL instance segmentation plays a pivotal role in various remote sensing applications, such as urban management, traffic planning, and agricultural estimation [5-7].

Attributed to the excellent feature learning capabilities, DL instance segmentation methods have shown significant promise in the remote sensing field. However, the intricacy of the background and the diversity of objects of interest make the RSI instance segmentation tasks still very challenging. Especially, large scenarios contained in RSIs call for long-range dependency modeling capabilities of DL models. Previous popular convolutional neural networks (CNNs), limited by the local receptive field of neurons, inherently struggle to handle object instances that are scattered across vast spatial regions but tightly semantic related [8]. Transformers inherently possess an excellent capacity to capture global contextual information, and have become popular in the RS community. The original ViT [9] and the more recent Swin Transformer [10] have achieved notable success in the computer vision field. However, ViT is limited by high computational requirements and low spatial resolution for feature extraction; Swin Transformer has certain limitations when applied to RSIs due to its local self-attention mechanism.

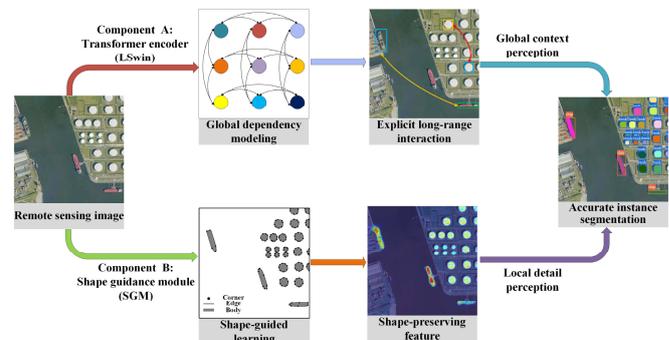

Fig. 1. The proposed instance segmentation method incorporates two key components that achieve local detail perception and global context perception, respectively. Component A is the new transformer encoder, and component B is the shape guidance module (SGM). They are respectively used to tackle the long-term dependency and precise boundary recognition problem in RSI instance segmentation tasks, aiming to achieve accurate instance shape delineation from broad geographical regions.

Previous studies in the RS community have tried pure transformer encoders [7], hybrid CNN-transformer encoders [11], and frozen transformer encoders transferred from foundational vision models [12] to extract better features. However, there is room for better long-range correlation modeling. For example, Swin Transformer and many derived encoder structures use only a single paradigm of self-attention. There is also a trade-off problem between robust long-range dependency modeling and computational efficiency. Swin Transformer employs a shifted window scheme, dividing the input images into multiple local windows and performing self-attention computation within each window. While this approach enhances efficiency, it limits self-attention

Manuscript was submitted on September 9, 2024. This work was supported by the National Natural Science Foundation of China under Grant 42171430 and Grant 42030102. (Corresponding author: Shunping Ji.)

D. Yu is with The Academy of Digital China, Fuzhou University, Fuzhou 350108, China (e-mail: yudawen@fzu.edu.cn).

S. Ji is with the School of Remote Sensing and Information Engineering, Wuhan University, Wuhan 430079, China (e-mail: jishunping@whu.edu.cn).

(correlation calculation) to local windows, resulting in insufficient global dependency modeling and potential cumulative long-range relationship errors as geospatial distance increases. In this paper, we introduce a global linear self-attention mechanism to local shifted-window based Swin Transformer and develop a novel transformer-style encoder, termed LSwin. The global linear self-attention mechanism conducts explicit long-range information interaction with acceptable GPU memory consumption.

Accurate shape delineation is the essential requirement of RSI instance segmentation. Shape guided learning in the computer vision field has introduced contour losses to facilitate the model to perceive object boundaries and learn the differences between the boundary of objects and the background [13]. Researchers in the RS community have also introduced specialized loss functions [1] [2] to enhance the object edges, which are usually regarded as hard-to-segment pixels. However, these methods mainly emphasize local boundaries while neglecting global contexts, resulting in a lack of holistic clues for adjacent or related object instances in RSIs. In this work, we designed an effective shape guidance module (SGM) that is more sensitive to object shapes and boundaries with specialized supervision information. Finally, we incorporate the local SGM and global LSwin in a unified network, as shown in Fig. 1, to complementarily capture local details and global contexts.

The main contributions of this study are summarized as follows:

1) We propose SGTN, a novel end-to-end model for RSI instance segmentation, which focuses on the two key issues, the long-term dependency and precise boundary recognition by incorporating a new transformer encoder and a shape guidance module.

2) The developed encoder offers superior global perception capability for large-scale RSIs, efficiently integrating mutual information to interpret related object instances scattered across broad spatial regions. The designed shape guidance module enhances boundary perception through specialized supervision, generating shape-preserving features and improving the quality of instance mask predictions.

## II. RELATED WORKS

### A. General Instance Segmentation

DL-based instance segmentation methods can be grouped into two types: pixel-segmentation based methods and contour-regression based methods. Pixel-segmentation based methods locate the objects of interest first and then segment their foreground region in a pixel-by-pixel manner, which include both single-stage methods, such as YOLACT [14], SOLO [15], BlendMask [16], CondInst [17]; and two-stage methods, like Mask R-CNN [18], Cascaded Mask R-CNN [19], HTC [20], PANet [21]. Contour-regression based methods directly predict the boundaries or sequential corners of objects. The popular contour-regression based methods include PolarMask [22], Curve-GCN [23], Deep Snake [24], DANCE [25], Polygon RNN [26], and Polygon RNN++ [27]. These general instance segmentation methods have been widely adopted as baseline models in the field of remote sensing, as detailed below.

### B. RSI Instance Segmentation

Researchers have proposed various strategies to improve these general instance segmentation methods in challenging scenarios in RSIs. Pixel-segmentation based methods have been widely explored in previous studies [1], [2], [5], [28], [29-31]. Su et al. [28] developed HQ-ISNet, introducing HRFPN [32] to fuse multi-level features and designing ISNetV2 to refine mask information flow between multiple mask prediction branches, which is based on Cascaded Mask R-CNN [33]. Feng et al. [29] developed SLCMASK-Net from Mask R-CNN, applying a sequence local context module to avoid confusion between closely situated instances. Yang et al. [3] proposed a cross-scale adaptive fusion module to aggregate multi-scale features and improve the detection and segmentation of objects of various sizes, based on PANet.

Contour-regression based methods have also been studied in the RS field. For example, Huang et al. [34] combined CNN with Recurrent Neural Networks (RNN) to produce closed building contours by sequentially predicting the corner points of buildings, with each building first located by Mask R-CNN. Similarly, Zhao et al. [35] applied RNN to decode the positions of corner points on the instance contour one by one, connecting them in order until a closed contour is obtained. Wei et al. [36] designed the two-stage BuildMapper to extract building instance contours. In the first stage, initial coarse contours were extracted, and in the second stage, a contour refinement procedure adjusted the contours to achieve regular polygon delineation.

### C. Long-range correlation modeling

One of the challenges in RSI instance segmentation lies in the tight relationships of objects scattering in a broad geographical region. The local receptive fields in standard convolution operations are naturally insufficient for global contextual information capture. Researchers in the RS field have explored and employed various methods for modeling long-range dependency between distant features or objects.

The CNN attention mechanisms [37-40] are exploited in the early years for long-range dependency modeling. Recently, transformers, such as ViT [9] and Swin Transformer [10], have become popular for modeling global dependencies. The computation in the multi-head self-attention (MSA) layers of a transformer block inherently models global dependencies, making it naturally adept at capturing long-range correlation relationships. ViT performs self-attention directly on all image tokens, while Swin Transformer achieves global feature correlation through a shifted-window based self-attention mechanism. ViT and Swin Transformer have been widely applied in semantic segmentation and object detection tasks [41], [42], but their implementation in RSI instance segmentation methods is still relatively limited. In previous studies, Xu et al. [11] combined the advantages of CNNs and transformers as the backbone for RSI instance segmentation. A more recent study by Chen et al. [12] proposed an RSI instance segmentation method evolved from the pre-trained Segment Anything Model [43], which uses the frozen ViT as the backbone. Zhong et al. [7], used a multi-scale Swin Transformer encoder called MSMTransformer.

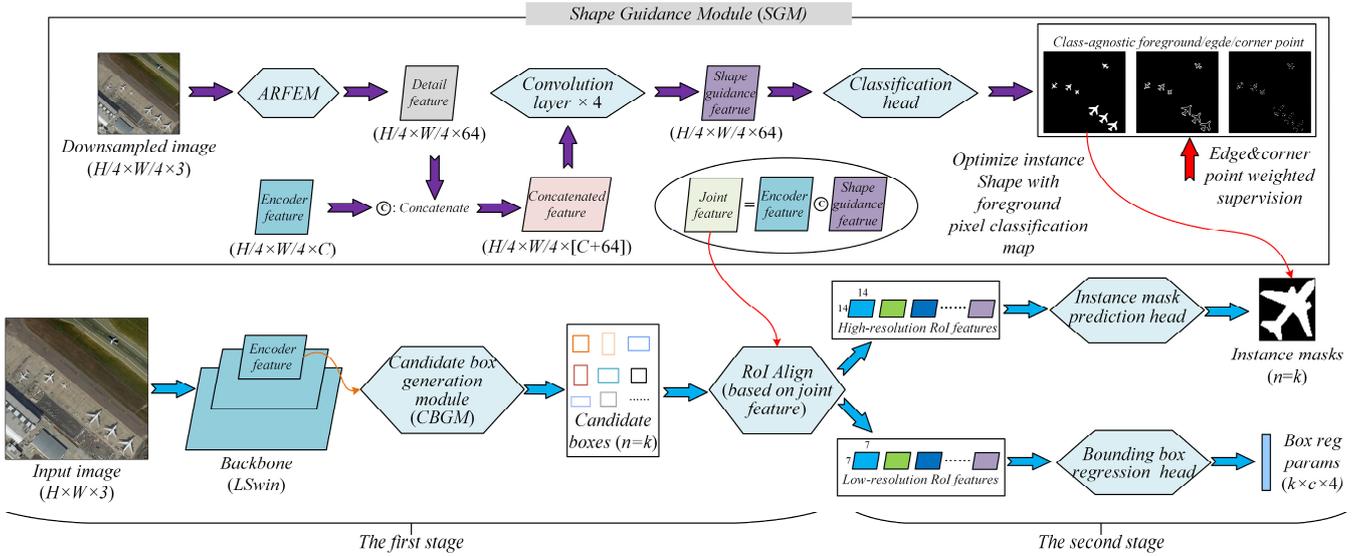

Fig.2. The network structure of our proposed SGTN. Candidate box generate module and ARFEM are from our previous works [8] [49], RoI Align, instance mask prediction head, and bounding box regression head are borrowed from Mask R-CNN. "W, H and C" *are width, height of the input image and feature channel number, and 'c' is the number of categories in the training dataset. In shape guidance module, a convolution layer = a 3×3 convolution + a ReLU + a Batch Normalization, classification head consists of two convolutional layers and a 1×1 convolution operation with three output channels.*

ViT conducts global self-attention operations across all tokens of the input image simultaneously, leading to significant GPU memory consumption. Swin Transformer reformulates the self-attention computation to local small windows to reduce computational complexity, but this also shortens the explicit correlation range. In this work, we propose a new transformer-style encoder devoted to RSI instance segmentation tasks. This encoder differs significantly from [7], [11], [12]. Our developed encoder couples the shifted-window-based self-attention paradigm of the Swin Transformer with a 1-D linear self-attention scheme. This linear self-attention sequentially computes feature correlations in 1-D vertical and horizontal directions, maintaining explicit long-range correlations with low GPU cost, which can compensate for the shortcomings of the Swin Transformer by providing sufficient global-perception capability for large-capacity RSIs.

### D. Shape guided learning

Shape or boundary is beneficial information for instance segmentation. In the computer vision field, shape guided learning has been utilized to enhance object contour extraction. For example, Takikawa et al., [44] and Chen et al., [45] adopted the multi-task network to predict additional object boundary maps, which can prompt the models to produce boundary-preserving features. Similar works can be found in [46], which embeds the boundary features generated through supervised learning into the instance mask prediction branch. Other studies emphasize more on the edge pixels than other pixels during training by assigning different pixels with different weight values [47], [48].

In the RS field, researchers have also paid attention to further utilizing the shape information by introducing specialized loss functions or boundary-aware modules. For example, Gong et al. [2] proposed an online hard sample mining strategy to enhance the network's attention to object edges, they assigned higher weights to pixels closer to the hard-to-segment instance boundaries. Wang et al. [1] explicitly modeled feature representations for the edges of objects by employing additional edge supervision.

In contrast to the above methods that only emphasize the local boundaries and the auxiliary boundary outputs are not explicitly used to improve the final instance segmentation masks, we propose the shape guidance module (SGM) along with the robust long-range correlation backbone to enrich the model with both local information and global contexts. The auxiliary output of SGM, i.e., the shape-aware foreground classification map, is also used to refine the instance masks by explicit computational rules.

## III. METHODS

### A. Overview of SGTN

The network structure of our proposed Shape Guided Transformer Network (SGTN) is shown in Fig. 2. SGTN adopts the popular two-stage framework, receiving remote sensing images as input and predicting first the locations (bounding boxes) and then the shapes of all instances in an end-to-end manner. The workflow of SGTN can be summarized as follows.

In the first stage, SGTN generates high-level semantic features and candidate bounding boxes. The feature encoder backbone of SGTN extracts discriminative features from the original images. To reduce discrepancies in features from different instances of the same category and describe various categories from a global perspective, a new transformer-style long-range correlation encoder is designed in this work. Sequentially, the anchor-free candidate box generation module (CBGM) predicts $N$ bounding boxes based on the encoded features. In this study, the center-based mechanism is adopted to locate the objects of interest, as in our previous work [49]. Parallel to CBGM, the shape guidance module (SGM) enhances the encoded features to accurately represent the shape details of objects by introducing prior supervision signals, specifically, the pixel-wise ground truth map of category-agnostic foreground, edges, and corner points.

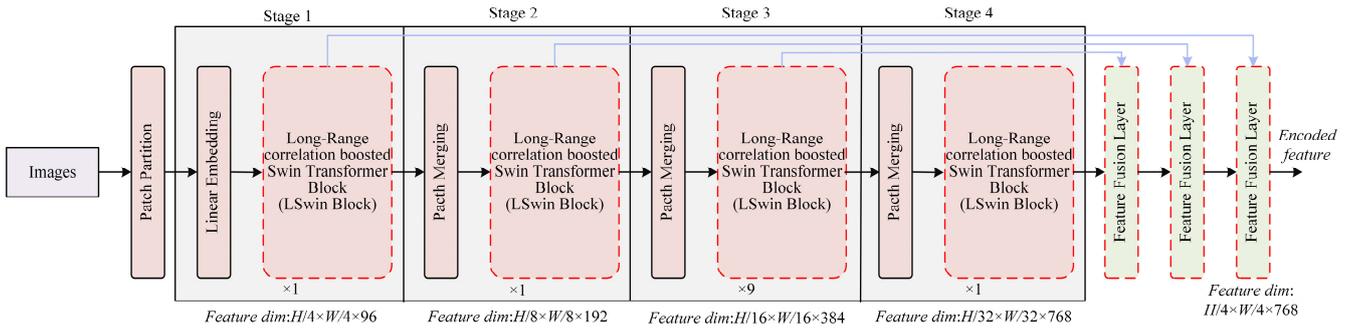

Fig. 3. The architecture of the Long-range correlation boosted Swin Transformer (LSwin) encoder. Components that differ from the original Swin Transformer are indicated with red dashed lines.

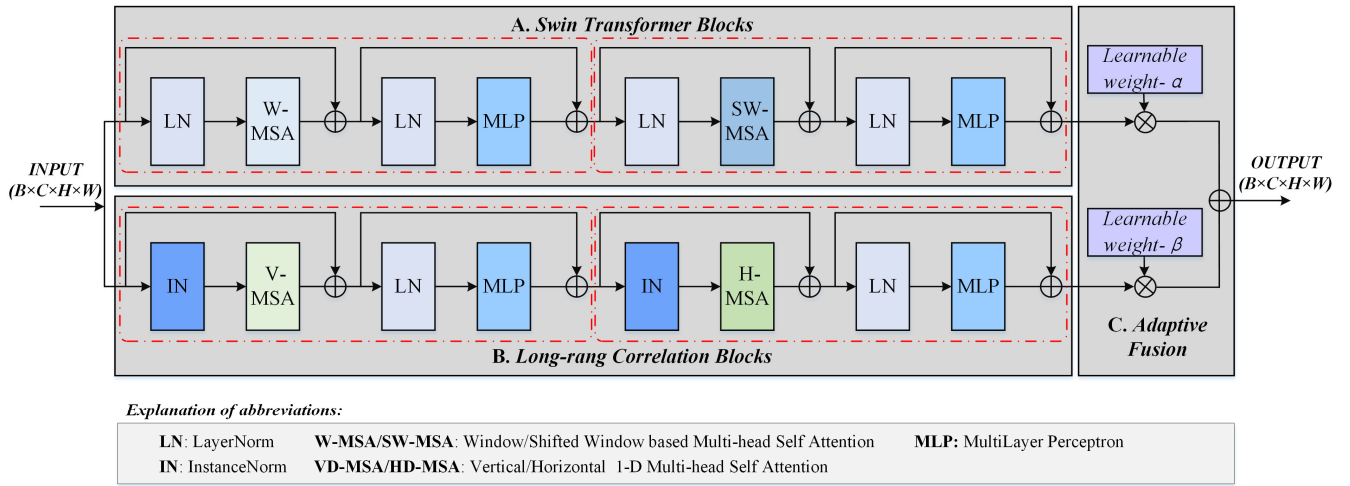

Fig. 4. The architecture of LSwin block. A shows two successive Swin Transformer blocks, highlighted in the top red dashed boxes. B illustrates two successive Long-range correlation blocks, highlighted in the bottom red dashed boxes. A LSwin block consists of A, B and C.

In the second stage, the joint features from SGM, and the candidate boxes from CBGM are used for instance mask prediction and bounding box refinement, similar to Mask R-CNN. The binary cross-entropy (BCE) and Dice loss functions are applied to the instance mask prediction branch [50], while the smooth L1 and CIoU loss functions are applied to the bounding box regression branch [49]. The fixed-size 28×28-pixel instance mask generated from the instance mask prediction branch is resized to align with the actual sizes in the input image, based on the regression-corrected bounding boxes. This resizing process may lead to the loss of spatial details. Inspired by previous research [5], the global-scale foreground classification map generated from SGM is used to optimize the resized instance masks and refine the instance shapes. First, we locate and crop the global-scale foreground classification map based on the bounding boxes to obtain the foreground mask of each individual instance, denoted as $M_c$. The resized instance mask, denoted as $M_s$, is fused with the corresponding $M_c$ using the Hadamard product to produce the final instance mask, referred to as $M_i$, i.e., $M_i = M_c \odot M_s$. In the following sections, we provide a detailed introduction to the newly proposed transformer-style long-range correlation encoder and shape guidance module.

### B. Long-range correlation boosted Swin Transformer encoder

In our previous work [8], we investigated a new paradigm of long-range feature correlation, in which the 2-D self-attention computation is decomposed into two 1-D correlations in orthogonal directions, i.e., vertical and horizontal directions. This new paradigm significantly reduces computational complexity while maintaining a long and explicit correlation range. To construct a feature encoder with outstanding global context encoding capability, we embedded the 1-D multi-head self-attention paradigm into Swin Transformer. The resulting encoder is named Long-range correlation boosted Swin Transformer, or LSwin for short.

The architecture of LSwin encoder is presented in Fig. 3, which is based on the small version of Swin Transformer

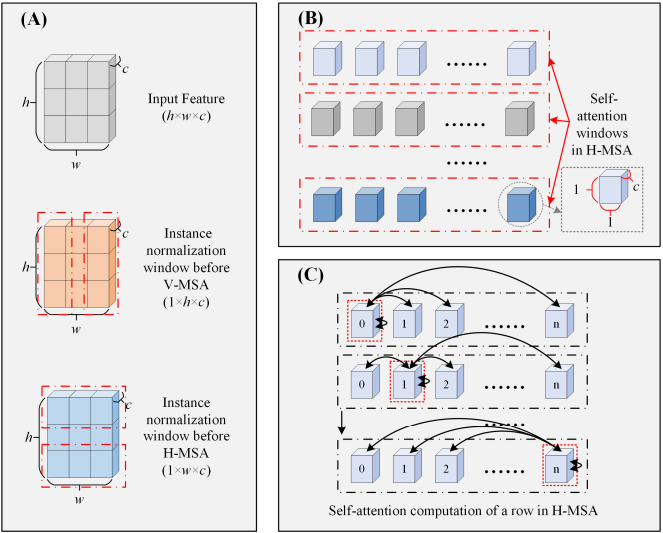

Fig. 5. Illustrations of the instance normalization window before V-MSA and H-MSA (A), the self-attention computation window in H-MSA (B), and the self-attention of a row in H-MSA (C).

(Swin-S) [10]. In the Patch Partition module of the encoder, an input RGB image is split into non-overlapping 4×4 patches, and all pixels in each patch are concatenated into a token, i.e., each token has a dimension of 4×4×3=48. These tokens are then projected to a given dimension (96 in our case study) by a linear embedding layer. Sequentially, several LSwin blocks, as shown in Fig. 4, are applied to further process these tokens. Patch merging operations at the beginning of stages 2 to 4 reduce the number of tokens and generate hierarchical feature representations. The feature produced at the end of stage 4 has a dimension of H/32×W/32×768, H, and W represent the height and width of the input image, respectively.

We further employ three feature fusion layers to improve the spatial resolution of the feature for subsequent bounding box regression and instance mask prediction procedures. The Patch Partition, linear embedding, and patch merging modules are the same as in Swin-S, and readers can refer to [10] for more details.

LSwin block is constructed by adding the long-range correlation blocks to the original Swin Transformer blocks. The long-range correlation blocks have a symmetric structure with Swin Transformer blocks. The computation of two successive Swin Transformer blocks in Fig. 4 can be detailed as follows:

$$\hat{z}_{sw}^l = \text{W - MSA}(\text{LN}(z^{l-1})) + z^{l-1}$$
$$z_{sw}^l = \text{MLP}(\text{LN}(\hat{z}_{sw}^l)) + \hat{z}_{sw}^l$$
$$\hat{z}_{sw}^{l+1} = \text{SW - MSA}(\text{LN}(z_{sw}^l)) + z_{sw}^l$$
$$z_{sw}^{l+1} = \text{MLP}(\text{LN}(\hat{z}_{sw}^{l+1})) + \hat{z}_{sw}^{l+1} \qquad (1)$$

where W-MSA/SW-MSA is the regular window/shifted window based self-attention operation, MLP means the multi-layer perceptron, and LN means the layer normalization. $Z^{l-1}$ denotes the input feature, $\hat{z}_{sw}^l / \hat{z}_{sw}^{l+1}$ represents the output feature of the W-MSA/SW-MSA at the Swin Transformer block of layer $l$/$l$+1, and $z_{sw}^l / z_{sw}^{l+1}$ is the output feature of MLP at the Swin Transformer block of layer $l$/$l$+1. Note that two successive blocks are different in Swin transformer, thus these blocks always appear in pairs.

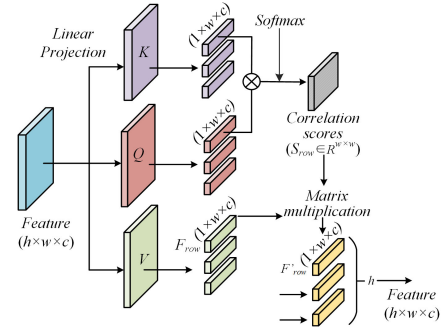

Fig. 6. Illustration of the self-attention operation in H-MSA.

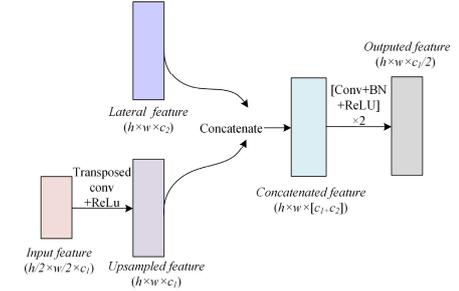

Fig. 7 The detailed structure of the feature fusion layer.

Similarly, the computations of two consecutive long-range correlation blocks in Fig. 4 can be summarized as follows:

$$\hat{z}_{lr}^l = \text{V - MSA}(\text{IN}(z^{l-1})) + z^{l-1}$$
$$z_{lr}^l = \text{MLP}(\text{LN}(\hat{z}_{lr}^l)) + \hat{z}_{lr}^l$$
$$\hat{z}_{lr}^{l+1} = \text{H - MSA}(\text{IN}(z_{lr}^l)) + z_{lr}^l$$
$$z_{lr}^{l+1} = \text{MLP}(\text{LN}(\hat{z}_{lr}^{l+1})) + \hat{z}_{lr}^{l+1} \qquad (2)$$

where V-MSA and H-MSA refer to the vertical and horizontal 1-D self-attention operation, respectively. We use instance normalization (IN) [51] to normalize the feature vector in the 1-D computation window of self-attention. $\hat{z}_{lr}^l / \hat{z}_{lr}^{l+1}$ represents the output feature of the V-MSA/H-MSA at the long-range correlation block of layer $l$/$l$+1, and $z_{lr}^l / z_{lr}^{l+1}$ is the output feature of MLP at the long-range correlation block of layer $l$/$l$+1.

The output features from two consecutive Swin Transformer blocks and long-range correlation blocks are combined using an adaptive fusion scheme. We employ two learnable parameters, $\alpha$ and $\beta$, to recalibrate the features from these two groups of blocks and then add them pixel-wise to produce the output feature of the LSwin block. In LSwin, Swin Transformer blocks are initialized with pre-trained weights from ImageNet, while the long-range correlation blocks are trained from scratch. Therefore, we set the initial value of $\alpha$ to 1 and $\beta$ to 0.

### C. Long-range correlation blocks

Details about the long-range correlation blocks are illustrated in Fig. 5 and Fig. 6. Given an input feature $F^{in} \in R^{h \times w \times c}$ of the long-range correlation blocks, $F^{in}$ is first split into $w$ columns and normalized before V-MSA. The instance normalization operation has a window size of $1 \times h \times c$. Here, $h$, $w$, and $c$ denote the height, width, and number of channels of the input feature, respectively. Similarly, the input feature is split into $h$ rows and normalized before H-MSA, and the instance normalization operation has a window size of $1 \times w \times c$.

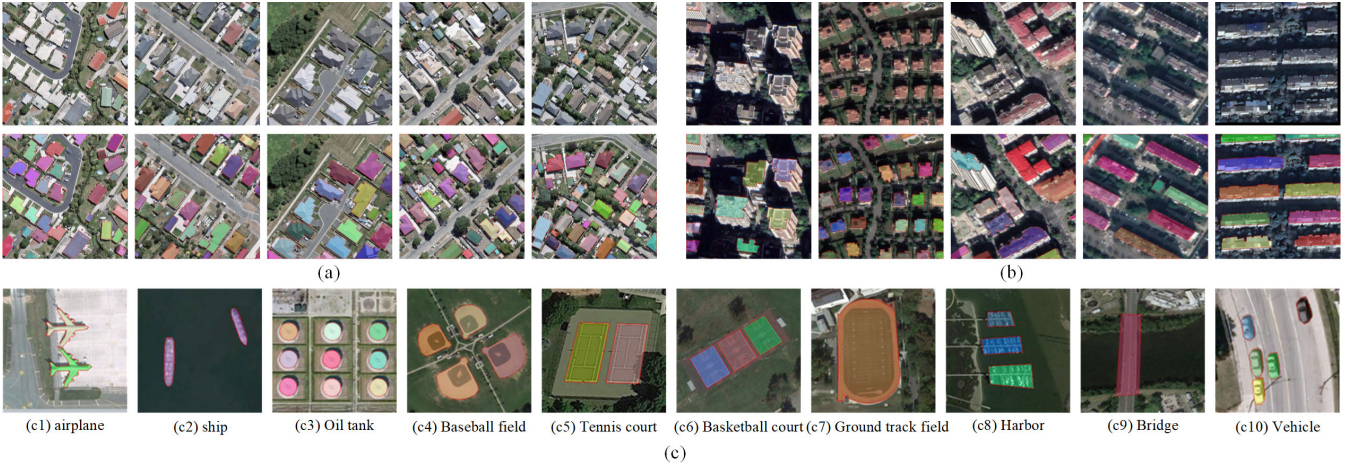

Fig. 8. Examples of remote sensing images and their corresponding instance annotations in (a) the WHU dataset, (b) the BITCC dataset, and (c) the NWPU VHR-10 dataset.

The normalized feature is then input into the V-MSA/H-MSA for self-attention operations along columns and rows, rather than performing feature correlations across all tokens of the entire input (such as ViT) or within a local 2-D window (such as Swin Transformer). Taking the computation process in H-MSA as an example, illustrated in Fig.5, its self-attention window has a dimension of $1 \times w$, with each token in it having a size of $c$. In H-MSA, three linear projection layers first map the normalized feature to three different feature vectors: $K_{row} \in R^{h \times w \times c}$ (key), $Q_{row} \in R^{h \times w \times c}$ (query), and $V_{row} \in R^{h \times w \times c}$ (value), respectively. The output feature of H-MSA is computed as a weighted sum of the $V_{row}$. The weights for each row of features in $V$ are computed using a compatibility function [52] of the feature (a row) in $Q_{row}$ with the corresponding feature (a row) in $K_{row}$. The calculation for a row of the output feature is as follows:

$$S_{row} = \text{Softmax}(F_{row}^Q F_{row}^{K^{\mathrm{T}}})$$
$$F_{row} = S_{row} F_{row}^V \tag{3}$$

where $F_{row}^V \in R^{1 \times w \times c}$, $F_{row}^Q \in R^{1 \times w \times c}$, $F_{row}^K \in R^{1 \times w \times c}$ represent a row of features in $V_{row}$, $Q_{row}$, and $K_{row}$, respectively. The superscript T denotes matrix transpose, $S_{row} \in R^{1 \times w \times w}$ represents the weights, and $F_{row} \in R^{1 \times w \times c}$ denotes a row of the output feature. Similarly, a column of the output feature from H-MSA is calculated as follows:

$$S_{col} = \text{Softmax}(F_{col}^Q F_{col}^{K^{\mathrm{T}}})$$
$$F_{col} = S_{col} F_{col}^V \tag{4}$$

where $F_{col}^V \in R^{1 \times h \times c}$, $F_{col}^Q \in R^{1 \times h \times c}$, $F_{col}^K \in R^{1 \times h \times c}$ represent a row of features in $V_{col}$, $Q_{col}$, and $K_{col}$, respectively. $S_{col} \in R^{1 \times h \times h}$ is the weight matrix, and $F_{col} \in R^{1 \times h \times c}$ denotes a column of the output feature from V-MSA.

### D. Feature Fusion layer

Three feature fusion layers (FFLs) located at the end of the proposed encoder progressively upsample the outputted feature map from stage 4 to produce a high-resolution feature map (with a 4× downsampling stride to the original image), as depicted in Fig. 2. Skip connections are employed within FFL to integrate features from multiple scales. Each FFL first upsamples the input feature using a transposed convolution with a stride of 2. It then concatenates the upsampled feature with the lateral feature via a skip connection and applies two convolutional layers to recalibrate the concatenated feature. In this study, a convolution layer is defined as a 3×3 convolution followed by Batch Normalization and ReLU activation. The detailed structure of the FFL is illustrated in Fig. 7.

### E. Shape guidance module

The designed Shape Guidance Module (SGM) produces 4× downsampled contour-aware features to enhance the performance of instance mask segmentation. First, an Adaptive Receptive Field Feature Extraction Module (ARFEM) [8] takes the 4× downsampled image as input and outputs a detail-rich shallow-layer feature with the same resolution. SGM then concatenates this shallow detail feature with the encoded feature extracted by the SLwin backbone and refines their combination using four convolutional layers. Edges and corner points are crucial elements in describing the shapes of objects. To explicitly enhance shape details in the features, we feed the shape guidance feature into a multi-label classification head to predict category-agnostic foreground pixels, edge pixels, and corner pixels. The classification head comprises two convolutional layers and a convolution operation with three output channels. Please note that as we emphasize the geometric shape information of objects in this module, we do not highlight semantic information and only perform foreground pixel classification that does not distinguish categories. In the multi-label classification task, a pixel can belong to multiple categories, each is managed by a special channel of the predictor. We apply three binary cross-entropy loss functions to supervise the three channels of the prediction result, with weight coefficients doubled/quadrupled for edge pixels/corner point pixels in all three binary cross-entropy loss functions. Inspired by the work of Shi et al. [5], we employ the class-agnostic foreground classification map to further refine the instance masks predicted from a global perspective, as introduced in III-A.

## IV. EXPERIMENTS AND RESULTS

In this section, we conduct comparison experiments to assess the effectiveness of the proposed SGTN. Three publicly available datasets were employed in this study, i.e., the WHU dataset [53], the BITCC dataset [54], and the NWPU VHR-10 dataset [28]. Details of the three datasets are presented in Section IV-A; the evaluation metrics are introduced in Section IV-B; the comparison results of the proposed SGTN with other recent instance segmentation methods are provided in Section IV-C; the effectiveness analysis of the newly designed LSwin encoder and the shape guidance module is presented in Section IV-D and IV-E, respectively. Discussions about the model efficiency and encoder construction are presented in Sections IV-D.

### A. Datasets

**1) WHU dataset.** The WHU dataset is a large-scale remote sensing building instance segmentation benchmark dataset [53], which consists of high-resolution aerial images and around 186,500 building instance annotations, collected from Christchurch, New Zealand. The dataset is divided into a training set (70% of the buildings), a validation set (10% of the buildings), and a test set (20% of the buildings). We resampled the original aerial images to a ground resolution of 0.2 m and cropped the images to 512×512-pixel tiles. The tiles that contain at least a building instance are used in the experiments. 9420/1537/3848 tiles were finally obtained for training/validation/testing.

**2) BITCC dataset.** The Building Instances of Typical Cities in China dataset (BITCC dataset for short) [54] consists of high-resolution satellite images and the corresponding building instance annotations, collected from Beijing, Shanghai, Wuhan, and Shenzhen, China. The size of original images in the BITCC dataset is 500×500 pixels. We padded images at the right and bottom edges with zero value to reproduce 512×512-pixel tiles to adapt to mainstream learning networks. The images and ground truth annotations of Beijing, Shanghai, and Wuhan were used in our experiments. Following the dataset publisher's partition rule, 80% of the buildings are used for training and the rest 20% of the buildings are used for testing. We finally obtained 3687 training tiles with 30,466 building instances and 799 test tiles with 6,403 building instances.

**3) NWPU VHR-10 dataset.** The NWPU VHR-10 dataset contains 650 remote sensing images and corresponding instance annotations collected from Google Earth and ISPRS Vaihingen datasets [28]. The image size ranges from 430×543 pixels to 1028×1728 pixels, with a ground resolution of 0.08-2 meters. A total of 10 categories were labeled, including airplanes (c1), ships (c2), oil tanks (c3), baseball fields (c4), tennis courts (c5), basketball courts (c6), ground track fields (c7), harbors (c8), bridges (c9), and vehicles (c10). Following the same data partitioning rules as the dataset publisher, 70% of the dataset was randomly selected as the training set and the remaining 30% as the testing set in our experiments. Due to the varying sizes of images in the dataset, the training images were cropped offline with a 25% overlap rate to form 800×800-pixel tiles. For images with a size of less than 800×800 pixels, we filled their right and bottom sides with all-zero values. During the test procedure, the original image was cropped into 800×800-pixel tiles at a 25% overlap rate in an online manner and then fed into the networks. The prediction results were merged by non-maximum suppression to produce the complete instance segmentation maps corresponding to the original large-size images.

Examples of images and the corresponding instance annotations from the WHU dataset, BITCC dataset, and NWPU VHR-10 dataset are shown in Fig. 8.

### B. Evaluation Metrics and Experimental Settings

**1) Evaluation Metrics:**
We evaluate the performance of different methods with the standard MS COCO measure [55]. The average precision (AP) at 10 different mask intersection over union (IoU) thresholds for instances of all sizes is taken as the main criterion, which can be computed as follows:

$$AP = \frac{\sum_{i=1}^{i=10} AP_{50+(i-1)\times5}}{10} \quad (5)$$

where $AP_i$ is the average precision (area under the precision-recall curve) when the mask IoU threshold to count true positive instances is set to $i\%$. The $AP_{50}$, $AP_{75}$, $AP_S$, $AP_M$, and $AP_L$ are also reported for comparing different methods comprehensively. The $AP_S$, $AP_M$, and $AP_L$ are counted by calculating the alone AP for small-sized instances (area < $32^2$ pixels), medium-sized instances ($32^2$ pixels ≤ area < $96^2$ pixels), and large-sized instances (area ≥ $96^2$ pixels), respectively.

**2) Implementation details:**
All experiments in this paper were conducted on a Windows PC equipped with an NVIDIA GeForce RTX 3090 24G GPU and an Intel Core i9-12900KF CPU. For a fair comparison, we re-implemented all the comparison methods, including the one-stage methods YOLACT [14] and SOLO [15], the two-stage methods Mask R-CNN [18], CenterMask [56], Deep Snake [24], DANCE [25], and BuildMapper [36] (BuildMapper only was used for the building instance segmentation datasets), using the same PyTorch deep learning framework. Various data augmentation strategies, including random horizontal and vertical flips, random color jittering, and random size scaling are applied at the training stages of all methods. The maximum learning epochs were set to 30, 60, and 100 for the WHU dataset, BITCC dataset, and NWPU VHR-10 dataset, respectively, based on the sample scales of the three datasets.

### C. Comparison with the state-of-the-art methods

**1) WHU dataset.** The quantitative results of the proposed SGTN and other comparison methods are listed in Table I. The comparison methods, i.e., YOLCAT, SOLO, Mask R-CNN, Deep Snake, Center-Mask, DANCE, and Buildmapper, all use ResNet-50 as their backbone. For the proposed SGTN, we report its performances with three different backbones: ResNet-50, Swin-S, and the developed LSwin. First, LSwin proves to be more effective than ResNet-50 and Swin-S for extracting building instances, improving the AP scores of SGTN by 2.2% and 0.5%, respectively, according to Table I. Compared to all other methods, SGTN achieves the best scores across all indicators. With the same backbone as other methods, SGTN outperforms the second-best CenterMask by 0.2% in the main indicator of AP. SGTN strives to generate shape-aware features, with the highest $AP_{75}$ score demonstrating the advantage of SGTN in delineating fine instance shapes.

TABLE I
Quantitative results of different instance segmentation methods on the WHU dataset.

| Method | Backbone | AP (%) | $AP_{50}$ (%) | $AP_{75}$ (%) | $AP_S$ (%) | $AP_M$ (%) | $AP_L$ (%) |
|---|---|---|---|---|---|---|---|
| YOLACT [14] | ResNet-50 | 60.9 | 87.8 | 71.3 | 40.3 | 71.8 | 49.6 |
| SOLO [15] | ResNet-50 | 63.9 | 89.1 | 73.9 | 41.4 | 75.9 | 54.0 |
| Mask R-CNN [18] | ResNet-50 | 65.3 | 90.0 | 77.1 | 45.9 | 75.5 | 60.6 |
| CenterMask [56] | ResNet-50 | 72.2 | 92.3 | 83.0 | 53.4 | 81.5 | 59.2 |
| Deep Snake [24] | ResNet-50 | 68.7 | 90.7 | 79.3 | 48.3 | 79.6 | 52.6 |
| DANCE [25] | ResNet-50 | 69.7 | 91.3 | 80.1 | 49.1 | 80.5 | 56.1 |
| BuildMapper [36] | ResNet-50 | 71.8 | 88.9 | 81.4 | 47.4 | 83.2 | 53.5 |
| SGTN (Ours) | ResNet-50 | 72.4 | 91.4 | 82.6 | 53.2 | 81.7 | 60.1 |
| SGTN (Ours) | Swin-S | 74.1 | 92.7 | 84.4 | 56.0 | 82.9 | 64.6 |
| SGTN (Ours) | LSwin | **74.6** | **93.2** | **84.7** | **57.3** | **83.4** | **67.0** |

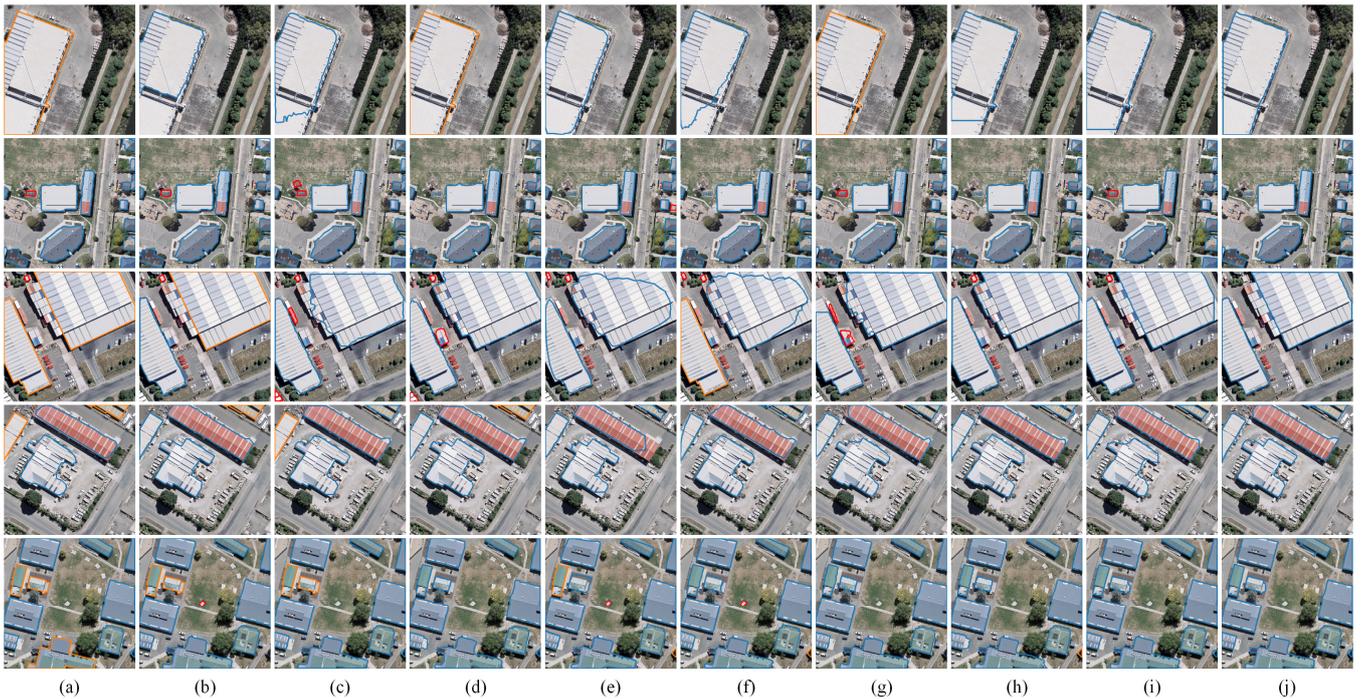

Fig. 9. The building instance segmentation results on the WHU dataset from different methods. The predicted true positive, false positive, and false negative instances in the results are colored in blue, red, and orange, respectively. (a)~(g) are the results from ResNet-50 based YOLACT, SOLO, Mask R-CNN, CenterMask, Deep Snake, DANCE, and BuildMapper, and (h)~(j) are the results from ResNet-50 based SGTN, Swin-S based SGTN, and LSwin based SGTN.

Fig. 9 shows several examples of building instance segmentation results on the WHU dataset from different methods. True positive instances, false positive instances, and false negative instances are annotated in blue, red, and orange, respectively. Compared to other methods, SGTN consistently shows fewer omissions and misclassifications. Furthermore, as shown in the last three columns of Fig. 9, the LSwin backbone intuitively enhances SGTN to depict buildings of varied sizes, colors, textures, and shapes more accurately.

*2) BITCC dataset.* The quantitative results from different instance segmentation methods on the BITCC dataset are listed

in Table II. Similar to the conclusions drawn from the WHU dataset, LSwin is highly effective for remote sensing instance segmentation tasks, providing significant performance improvements compared to ResNet-50 and Swin-S backbones. Additionally, SGTN proves more advantageous than other recent instance segmentation methods, achieving the highest scores across all indicators. When using the same ResNet-50 backbone, SGTN outperforms the second-best CenterMask by 1.0% in terms of AP. When using our LSwin backbone, SGTN exceeds 2.4 AP. The significant advantage of SGTN in the $AP_{75}$ indicator also demonstrates that the newly developed SGM significantly enhances the model's ability to depict detailed shapes of building instances.

Examples of building instance segmentation results on the BITCC dataset from different methods are shown in Fig. 10. True positive instances (blue contours), false positive instances (red contours), and false negative instances (orange contours) are annotated in the original images. The BITCC dataset presents considerable challenges due to oblique shooting perspectives, low spatial resolution, poor imaging conditions, and occasional indistinguishability between buildings and backgrounds. All comparison methods show relatively lower instance segmentation quality for building boundaries compared to the WHU dataset. However, SGTN consistently results in fewer false positives and false negatives than other methods when using ResNet-50 as the backbone. The advantage of the newly designed LSwin backbone is also evident in the last three columns of Fig. 10, where LSwin-based SGTN effectively segments building instances of various styles.

TABLE II
QUANTITATIVE RESULT OF DIFFERENT INSTANCE SEGMENTATION METHODS ON THE BITCC DATASET.

| Method | Backbone | AP (%) | $AP_{50}$ (%) | $AP_{75}$ (%) | $AP_S$ (%) | $AP_M$ (%) | $AP_L$ (%) |
|---|---|---|---|---|---|---|---|
| YOLACT [13] | ResNet-50 | 40.4 | 67.9 | 43.0 | 17.4 | 46.3 | 38.0 |
| SOLO [14] | ResNet-50 | 42.0 | 71.5 | 45.1 | 17.0 | 46.3 | 41.3 |
| Mask R-CNN [17] | ResNet-50 | 43.1 | 67.6 | 49.1 | 19.0 | 46.8 | 48.2 |
| CenterMask [49] | ResNet-50 | 48.4 | 75.3 | 55.4 | 25.6 | 52.2 | 49.2 |
| Deep Snake [23] | ResNet-50 | 45.2 | 73.7 | 50.4 | 23.9 | 49.2 | 45.0 |
| DANCE [24] | ResNet-50 | 46.8 | 74.7 | 52.4 | 25.5 | 50.7 | 46.7 |
| BuildMapper [34] | ResNet-50 | 46.5 | 69.7 | 52.9 | 21.2 | 50.6 | 46.0 |
| SGTN (Ours) | ResNet-50 | 49.4 | 75.6 | 56.2 | 28.4 | 52.8 | 50.4 |
| SGTN (Ours) | Swin-S | 50.7 | 76.7 | 57.5 | 29.6 | 54.6 | 51.3 |
| SGTN (Ours) | LSwin | **52.1** | **78.4** | **59.8** | **31.6** | **55.1** | **54.2** |

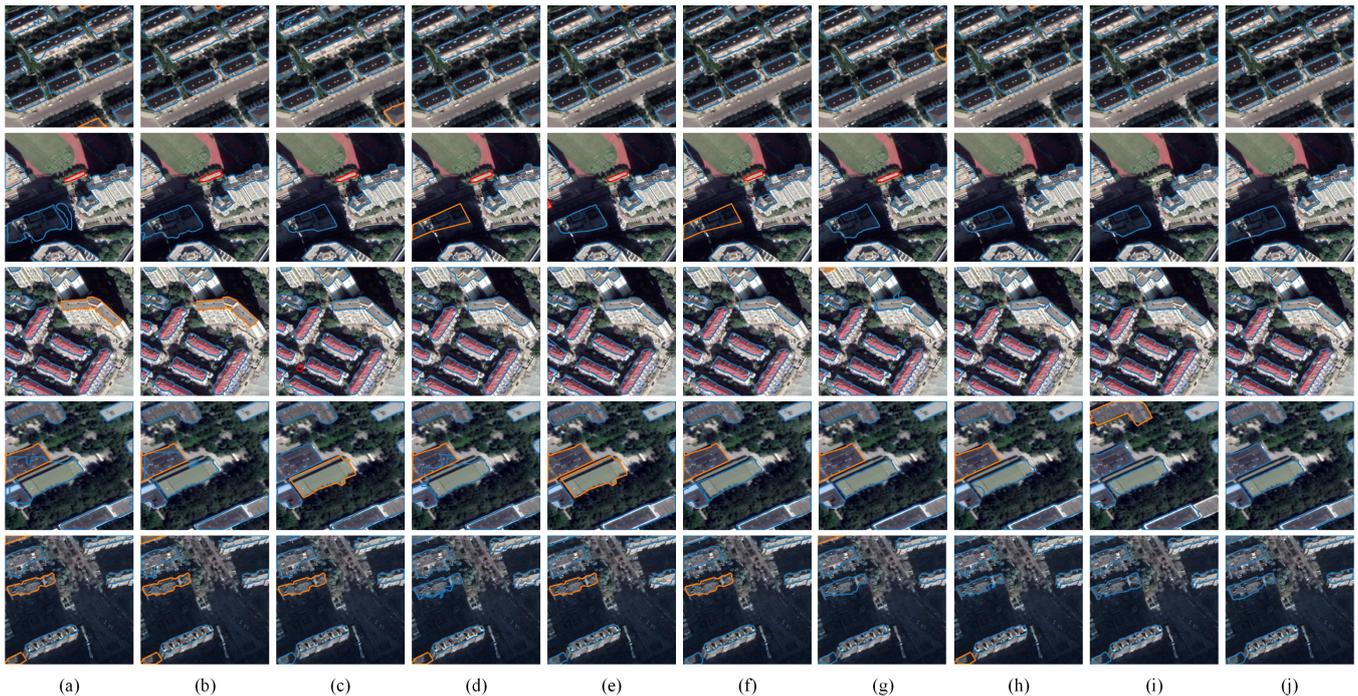

Fig. 10. The building instance segmentation results on the BITCC dataset from different methods. The predicted true positive, false positive, and false negative instances in the results are colored in blue, red, and orange, respectively. (a)–(g) are the results from ResNet-50 based YOLACT, SOLO, Mask R-CNN, CenterMask, Deep Snake, DANCE, and BuildMapper, and (h)–(j) are the results from ResNet-50 based SGTN, Swin-S based SGTN, and LSwin based SGTN.

*3) NWPU VHR-10 dataset.* Results on the multi-category NWPU VHR-10 dataset from different instance segmentation methods are summarized in Table III. SGTN achieves an AP of 69.3%, surpassing the second-best CenterMask by 2.8% when using ResNet-50 as the backbone. This highlights SGTN's reliability and adaptability in processing various types of objects of interest in remote-sensing images. Fig. 11 presents the AP scores for each category on the NWPU VHR-10 dataset, where SGTN excels with the highest AP scores in seven categories: airplane (c1), ship (c2), oil tank (c3), baseball court (c4), tennis court (c5), basketball court(c6), and vehicle (c10). Other methods only lead in specific categories: DANCE in ground track field (c7), Mask R-CMM in the harbor (c8), and CenterMask in bridge (c9). LSwin, as the designed backbone, demonstrates a 2.0% and 1.6% improvement in AP score compared to ResNet-50 and Swin-S, respectively, as shown in the last three rows of Table III.

To comprehensively compare SGTN's performance with other instance segmentation methods from the remote sensing field, we conducted experiments with more backbones, as shown in Fig. 12. Methods from Su et al. [57] (2019), Yang et al. [3] (2021), Wang et al. [1] (2022), Shi and Zhang [5] (2022), Su et al. [31] (2022), Chen et al. [12] (2024), and Chen et al. [58] (2024) were evaluated. Notably, when using the same backbone, SGTN consistently achieves higher AP scores compared to other methods. Moreover, LSwin, developed in this study, exhibits significant advantages when compared across different backbones. Fig. 12 clearly illustrates SGTN's

TABLE III
QUANTITATIVE RESULT OF DIFFERENT INSTANCE SEGMENTATION METHODS ON THE NWPU-VHR DATASET.

| Method | Backbone | AP (%) | AP$_{50}$ (%) | AP$_{75}$ (%) | AP$_S$ (%) | AP$_M$ (%) | AP$_L$ (%) |
|---|---|---|---|---|---|---|---|
| YOLACT [13] | ResNet-50 | 56.6 | 87.8 | 57.8 | 41.6 | 55.9 | 56.7 |
| SOLO [14] | ResNet-50 | 57.7 | 86.2 | 61.1 | 44.9 | 55.4 | 61.1 |
| Mask R-CNN [17] | ResNet-50 | 64.3 | 88.2 | 72.3 | 55.5 | 63.0 | 67.7 |
| CenterMask [49] | ResNet-50 | 66.5 | 92.0 | 72.4 | 53.9 | 66.0 | 66.5 |
| Deep Snake [23] | ResNet-50 | 63.7 | 90.2 | 67.9 | 51.9 | 62.7 | 62.0 |
| DANCE [24] | ResNet-50 | 66.3 | 90.8 | 74.2 | 55.1 | 65.2 | 67.5 |
| SGTN (Ours) | ResNet-50 | 69.3 | 91.7 | 77.7 | **59.1** | 69.2 | 67.3 |
| SGTN (Ours) | Swin-S | 69.7 | 91.9 | 76.9 | 57.2 | 69.3 | 68.2 |
| SGTN (Ours) | LSwin | **71.3** | **93.3** | **79.9** | 57.5 | **71.0** | **71.2** |

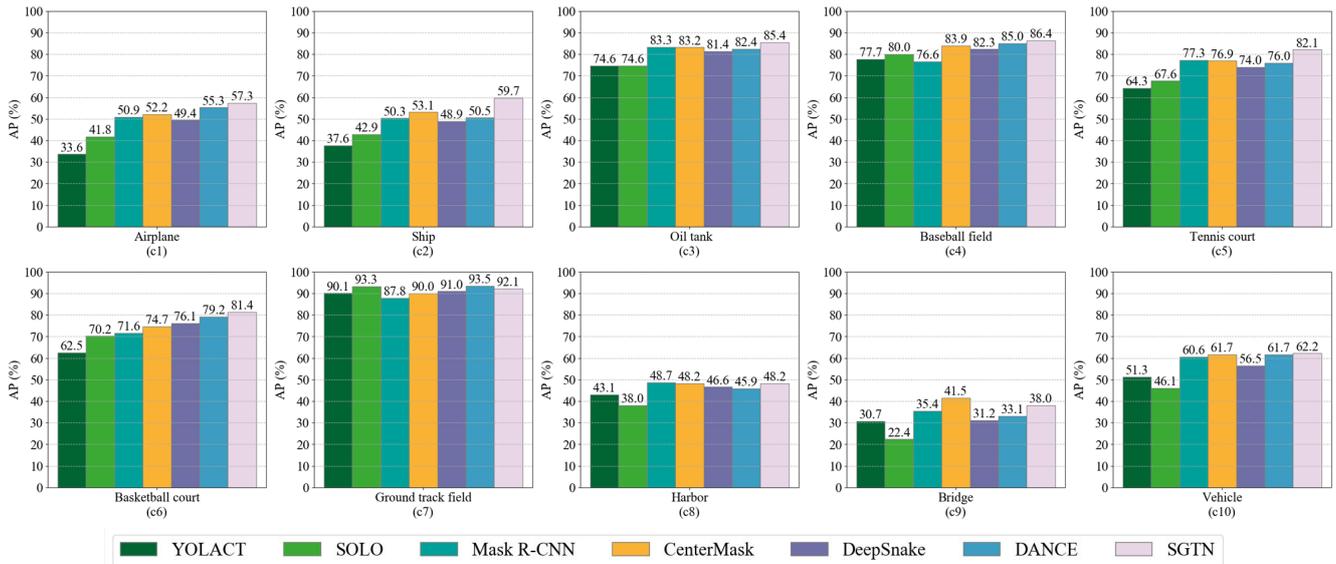

Fig. 11. The performance comparison of different methods on the NWPU VHR-10 dataset. (c1) to (c10) show the AP scores for each category on the NWPU VHR-10 dataset. ResNet-50 is used as the backbone for all the methods.

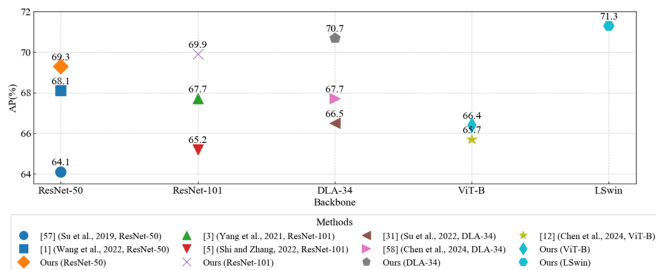

Fig. 12. The performance comparison between different instance segmentation methods from the remote sensing field.

superior performance over previous methods in the remote sensing field.

Visualization results of SGTN on the NWPU VHR-10 dataset are depicted in Fig. 13. True positive instances are annotated with colored bounding boxes and their corresponding category texts, with the same color indicating the same category. False negative and false positive instances are marked with red bounding boxes labeled "FN" and "FP," respectively. SGTN effectively locates and segments multi-class objects with varied shapes, orientations, sizes, colors, and textures, as evidenced in the first three rows of Fig. 13. While generally accurate, minor errors are observed, mainly due to similar shapes contributing to false positive segmentations, as seen in the first three columns of the last row. Additionally, specific challenges such as bridges connected to road networks, ships docked at the dock, and uniquely shaped harbors contribute to missed segmentations. SGTN demonstrates robust performance in handling diverse and challenging RSI instance segmentation tasks.

### D. Effectiveness of the new LSwin backbone

In this section, we conduct experiments to assess the effectiveness of combining the developed LSwin backbone with other instance segmentation methods. Specifically, CenterMask and DANCE, representing pixel-segmentation and contour-regression based methods respectively, are selected. The experimental results for CenterMask and DANCE are illustrated in Fig. 14. Using ResNet-50 as the baseline, the Swin-S improves the performance of CenterMask and DANCE on the BITCC and NWPU VHR-10 datasets, but results in decreased AP scores for DANCE on the WHU dataset. Our LSwin consistently enhances the performance of CenterMask and DANCE across all three datasets. Quantitatively, LSwin boosts the AP scores of CenterMask by 1.2%, 2.6%, and 3.1% on the WHU, BITCC, and NWPU VHR-10 datasets, respectively. For DANCE, the improvements of 0.3%, 2.4%, and 1.5% in AP scores are observed across these datasets. These significant enhancements in AP underscore the robust generality and scalability of LSwin, making it adaptable for integration with other advanced instance segmentation methods.

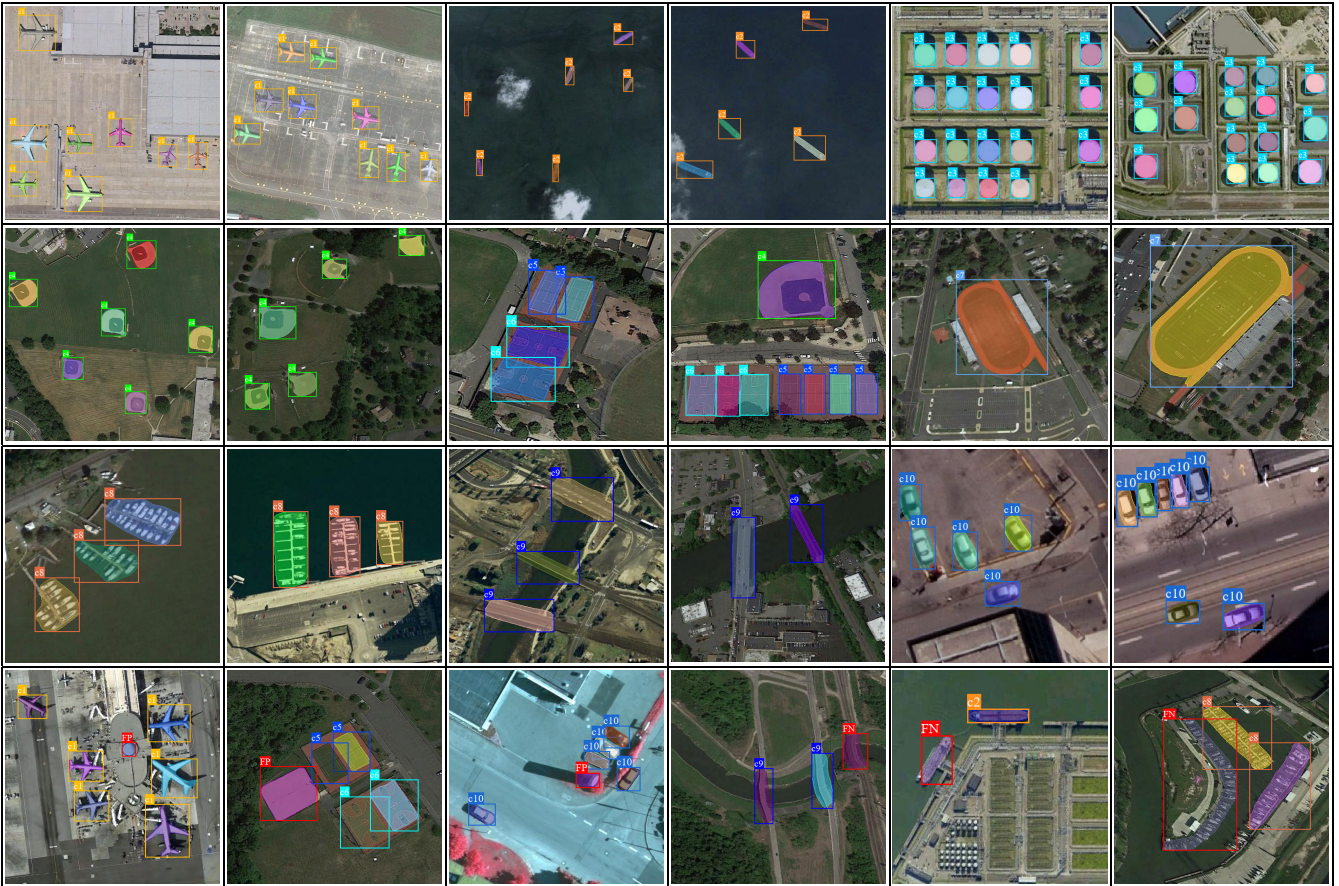

Fig. 13. The instance segmentation results on the NWPU VHR-10 dataset from SG-LRN. The predicted true positive instances are annotated with randomly colored bounding boxes attached with their category texts. The predicted false negative and false positive instances are annotated with red colored bounding boxes attached with the texts "FN" and "FP", respectively.

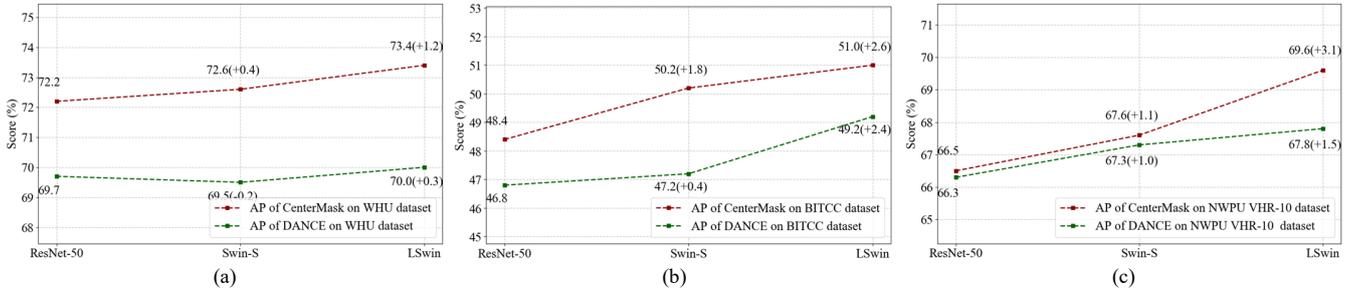

Fig. 14. The performance comparison of three different feature encoders on the WHU dataset (a), BITCC dataset (b), and NWPU VHR-10 dataset (c).

### E. Effectiveness of the shape guidance module

In our study, we introduced the shape guidance module (SGM) to produce the shape-preserving feature representation of the instance segmentation network using explicit supervision. The class-agnostic foreground classification map, i.e., an auxiliary output of SGM, is also utilized to refine the instance masks from a global perspective. In this section, we present ablation experiments where SGM is the variable under study, evaluating its effectiveness.

Table IV summarizes the results of the ablation study across the WHU dataset, BITCC dataset, and NWPU VHR-10 dataset. SGM improves the AP scores by 1.3%, 1.1%, and 1.4% for the three datasets, respectively. This improvement demonstrates SGM's capability to enhance instance segmentation performance, alleviating detail losses when predicting the instance masks at the end of the network. Notably, the consistent gains in $AP_{75}$ scores by SGM, i.e., 0.8% for the WHU dataset, 0.9% for the BITCC dataset, and 2.2% for the NWPU VHR-10 dataset, further underscore SGM's effectiveness in improving the fine shape prediction of our instance segmentation method.

Taking the WHU dataset as an example, the qualitative experimental results are presented in Fig. 15. The IoU scores between the predicted instance masks and the ground truth instance masks are marked on the predicted bounding boxes. SGM intuitively improves instance mask segmentation

Table IV
Ablation studies for the developed shape guidance module. "w/" denotes with, "w/o" denotes without.

| Dataset | Description | AP (%) | AP$_{50}$ (%) | AP$_{75}$ (%) |
|---|---|---|---|---|
| WHU dataset | SGTN w/ SGM | 74.6 | 93.2 | 84.7(+0.8) |
| | SGTN w/o SGM | 73.3 | 93.1 | 83.9 |
| BITCC dataset | SGTN w/ SGM | 52.1 | 78.4 | 59.8(+0.9) |
| | SGTN w/o SGM | 51.0 | 77.5 | 57.9 |
| NWPU VHR-10 | SGTN w/ SGM | 71.3 | 93.3 | 79.9(+2.1) |
| | SGTN w/o SGM | 69.6 | 93.7 | 77.8 |

Table V
Efficiencies of different methods for instance segmentation on the BITCC dataset.

| Method | Backbone | AP(%) | Run Time |
|---|---|---|---|
| YOLACT | ResNet-50 | 40.4 | 55.8sec |
| SOLO | ResNet-50 | 42.0 | 58.3sec |
| Mask R-CNN | ResNet-50 | 43.1 | 61.7sec |
| Center-Mask | ResNet-50 | 48.4 | 51.2sec |
| Deep Snake | ResNet-50 | 45.2 | 51.3sec |
| DANCE | ResNet-50 | 46.8 | 49.8sec |
| BuildMapper | ResNet-50 | 46.5 | 49.8sec |
| SGTN | ResNet-50 | 49.4 | 52.6sec |
| SGTN | Swin-S | 50.7 | 56.0sec |
| SGTN | LSwin | 52.1 | 66.3sec |

Table VI
Comparison experiments for different encoders.

| Dataset | Backbone | AP (%) | AP$_{50}$ (%) | AP$_{75}$ (%) |
|---|---|---|---|---|
| WHU dataset | Swin-S | 74.1 | 92.7 | 84.4 |
| | LRC | 71.2 | 90.7 | 81.2 |
| | LSwin | **74.6** | **93.2** | **84.7** |
| BITCC dataset | Swin-S | 50.7 | 76.7 | 57.5 |
| | LRC | 40.6 | 66.5 | 44.4 |
| | LSwin | **52.1** | **78.4** | **59.8** |
| NWPU VHR-10 | Swin-S | 69.7 | 91.9 | 76.9 |
| | LRC | 63.0 | 87.0 | 68.9 |
| | LSwin | **71.3** | **93.3** | **79.9** |

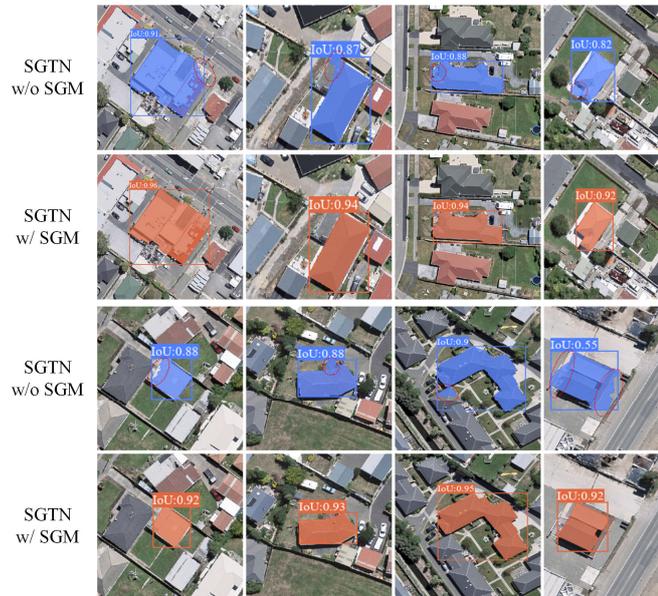

SGTN w/o SGM

SGTN w/ SGM

SGTN w/o SGM

SGTN w/ SGM

Fig. 15. Visual comparison of prediction results with and without embedding SGM in the proposed method.

accuracy, delineating the true shape and boundary position of the objects of interest close to the level of manual drawing. The improved IoU scores, as illustrated in Fig. 15, also indicate the effectiveness of SGM in enhancing the prediction quality of instance shapes.

### F. Discussion

**1) Inference Efficiency.** In this subsection, we analyze the inference efficiency of our newly developed method for instance segmentation of remote sensing images. We conducted this analysis using the multi-category BITCC dataset, and the runtime statistics of all comparison methods are summarized in Table V. Overall, the speeds of different methods are comparable. When employing ResNet-50 as the backbone, our proposed SGTN shows faster inference times compared to YOLACT, SOLO, and Mask R-CNN, while slightly slower than Center-Mask, Deep Snake, DANCE, and BuildMapper. Despite slightly lower efficiency, SGTN achieves the highest AP score of 49.4% with ResNet-50, outperforming other methods by at least 1.0%.

Another notable observation from Table V is that using Swin-S and LSwin as backbones is almost equally efficient compared to the classic ResNet-50, yet they also yield improvements in AP scores. This indicates the prospect of small-scale transformer backbones in remote sensing instance segmentation applications.

**2) Building an encoder from only long-range correction**

**blocks.** In this work, we build a new encoder by combining Swin Transformer blocks and efficient 1D Long-Range Correction (LRC) blocks. In this subsection, we discuss the effectiveness of using single LRC blocks within the encoder. We term the modified version of the encoder "LRC" and compared it with the original Swin Transformer encoder. We embed the LRC and Swin-S encoders into the proposed method respectively, and the corresponding experimental results are reported in Table VI.

The first observation is that the combination of Swin-S and LRC (i.e., LSwin) produces higher AP scores than using Swin-S or LRC alone across all three datasets. Notably, LRC alone performs significantly worse than Swin-S, showing a 10.1% decrease in AP score on the BITCC dataset, suggesting that LRC may not be suitable as a standalone feature encoder. Therefore, we combined LRC with Swin-S to build a robust feature encoder for remote sensing image instance segmentation tasks.

## V. Conclusion

This paper proposes a novel deep learning instance segmentation method, termed SGTN, which excels in global dependency modeling and local fine-grain perception, achieving precise instance shape delineation in remote sensing images. First, a long-range-correlation boosted Swin Transformer encoder (LSwin) is designed, innovatively integrating two different self-attention mechanisms. LSwin has demonstrated better global-perception capacity to large-capacity remote sensing images compared to the popular Swin Transformer and ResNet. Second, we develop the shape guidance module (SGM) to emphasize the shape detail of objects of interest by recalibrating the internal features with explicit supervision and producing the shape-preserving feature representation. Qualitative and quantitative results on three public single-class and multi-class remote sensing image datasets confirm that SGTN outperforms other recent instance segmentation methods. We have also conducted comprehensive ablation experiments to validate the effectiveness of LSwin and SGM. We hope the study presented in this paper will advance the research on deep learning-based instance segmentation methods in the field of remote sensing.